\documentclass[letterpaper, 10 pt, conference]{ieeeconf}  %

\usepackage[T1]{fontenc} %
\usepackage{graphicx}
\usepackage{hyperref}
\usepackage{amsmath} 

\usepackage{booktabs}
\usepackage{threeparttable}
\usepackage{pifont}
\newcommand{\cmark}{\ding{51}} %
\newcommand{\xmark}{\ding{55}} %
\usepackage{mathtools}
\usepackage{booktabs}
\usepackage[table]{xcolor}
\usepackage{multirow}
\usepackage{makecell}
\usepackage{geometry}
\usepackage{tabularx} %
\definecolor{headblue}{RGB}{210,230,255}
\definecolor{ioupink}{RGB}{255,225,235}
\definecolor{metricgreen}{RGB}{220,245,230}
\definecolor{methodgray}{RGB}{240,240,240}
\definecolor{groupbg}{RGB}{255,235,200}
\usepackage{hyperref}

\IEEEoverridecommandlockouts                              %

\title{\LARGE \bf
VLA-R1: Enhancing Reasoning in Vision-Language-Action Models
\vspace{-0.5cm}
}

\author{
    \textbf{Angen Ye}$^{12*}$\quad
    \textbf{Zeyu Zhang}$^{1*}$\quad 
    \textbf{Boyuan Wang}$^{12}$\quad 
    \textbf{Xiaofeng Wang}$^{13}$\quad
    \textbf{Dapeng Zhang}$^{2}$\quad 
    \textbf{Zheng Zhu}$^{1\dag}$ \vspace{0.1cm}\\
    $^1$GigaAI\quad
    $^2$CASIA\quad
    $^3$Tsinghua University\vspace{0.05cm}\\
    \small $^*$Equal contribution. $^\dag$Corresponding author: zhengzhu@ieee.org.
}

\begin{document}

\maketitle
\thispagestyle{empty}
\pagestyle{empty}

\begin{abstract}
Vision-Language-Action (VLA) models aim to unify perception, language understanding, and action generation, offering strong cross-task and cross-scene generalization with broad impact on embodied AI. However, current VLA models often lack explicit step-by-step reasoning, instead emitting final actions without considering affordance constraints or geometric relations. Their post-training pipelines also rarely reinforce reasoning quality, relying primarily on supervised fine-tuning with weak reward design. To address these challenges, we present \textbf{VLA-R1}, a reasoning-enhanced VLA that integrates Reinforcement Learning from Verifiable Rewards (RLVR) with Group Relative Policy Optimization (GRPO) to systematically optimize both reasoning and execution. Specifically, we design an RLVR-based post-training strategy with verifiable rewards for region alignment, trajectory consistency, and output formatting, thereby strengthening reasoning robustness and execution accuracy. Moreover, we develop \textbf{VLA-CoT-13K}, a high-quality dataset that provides chain-of-thought supervision explicitly aligned with affordance and trajectory annotations.  Furthermore, extensive evaluations on in-domain, out-of-domain, simulation, and real-robot platforms demonstrate that VLA-R1 achieves superior generalization and real-world performance compared to prior VLA methods. We plan to release the model, code, and dataset following the publication of this work.
Code: \url{https://github.com/GigaAI-research/VLA-R1}.
Website: \url{https://gigaai-research.github.io/VLA-R1}.
\end{abstract}

\section{Introduction}

\begin{figure}[htbp]
    \centering
    \includegraphics[width=\linewidth]{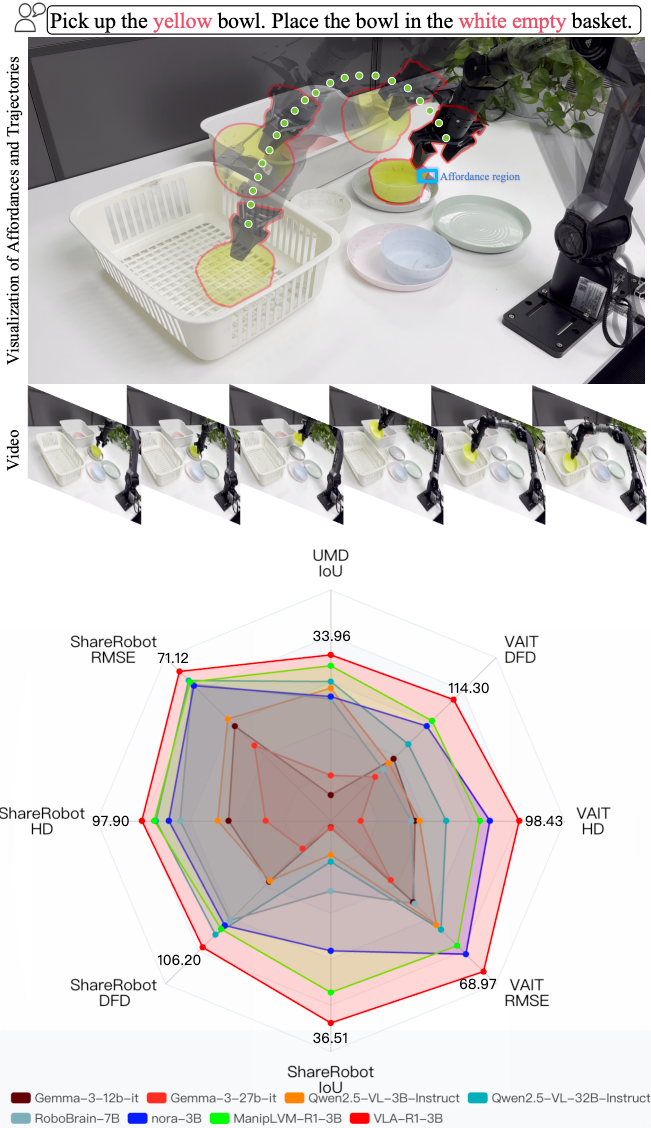}
    \caption{\raggedright \textbf{VLA-R1:} pipeline from instruction to execution, with benchmark comparisons against baselines.}
    \vspace{-2.2em}
    \label{fig:teaser}
    
\end{figure}

Vision–Language–Action (VLA) models unify perception, language, and action. They first learn open-vocabulary semantics and cross-modal alignment from internet-scale image–text pretraining. These semantics are then grounded into the action space through multi-task manipulation data. This enables analogical transfer to unseen objects and compositional generalization to novel commands. Compared with modular pipelines \cite{garrett2021integrated,ye2024non} or state-driven policies \cite{gu2017deep}, VLAs show stronger cross-task and cross-scene generalization \cite{song2025maniplvm, ji2025, kim2024,liu2024rdt, intelligence2504pi0,wang2025embodiedreamer}. Representative works include VoxPoser \cite{huang2023a} for zero-shot trajectory planning, and ManipLVM-R1 \cite{song2025maniplvm} and RoboBrain \cite{ji2025} for integrating affordance perception and pose estimation. Meanwhile, Reinforcement Learning from Verifiable Rewards (RLVR) enhances reasoning and generalization in vision–language models. Vision-R1 \cite{huang2025a} matches larger models through cold-start data and progressive training; LMM-R1 \cite{peng2025} employs a two-stage regimen from textual reasoning to multimodal tasks; and VLM-R1 \cite{shen2025} applies R1-style reinforcement to visual grounding, boosting open-vocabulary detection.

However, existing VLA models present two significant challenges. First, they often lack step-by-step reasoning: models tend to emit final actions directly without explicit inference over affordance constraints, geometric relations, or container selection. This limitation leads to instruction-disambiguation failures under color similarity, duplicate instances, or multiple candidate receptacles. 
Second, post-training rarely provides systematic reinforcement of reasoning. Current method relies on supervised fine-tuning (SFT) with little reward optimization targeted at reasoning quality and execution efficacy. Even when Reinforcement Learning (RL) is used, reward design is typically single-objective and struggles to jointly optimize region alignment and trajectory consistency, degrading performance on out-of-distribution data and in the real world.

To address these challenges, we propose \textbf{VLA-R1}, a post-training-enhanced VLA model capable of step-by-step reasoning. Unlike prior approaches, VLA-R1 simultaneously emphasizes data-level Chain-of-Thought (CoT) supervision and optimization-level reward alignment, bridging the gap between reasoning and execution. This enables the model to not only provide answers but also explain them, making it robust to challenges like color similarity, repeated instances, and multiple receptacle choices during reasoning.

To further enhance the model's reasoning capabilities, we introduce an RLVR-based post-training strategy at the optimization layer. Specifically, we employ Group Relative Policy Optimization (GRPO) \cite{shao2024deepseekmath} with three verifiable rewards: an affordance reward based on Generalized Intersection over Union (GIoU) \cite{rezatofighi2019generalized} to provide informative gradients for non-overlapping predicted and ground truth affordance regions, speeding up learning; a distance-based reward using the improved Fréchet distance to ensure reasonable trajectory curvature and segment length; and an output-format reward to enforce well-formed reasoning and action specifications. These optimizations enable VLA-R1 to generate accurate affordance regions and well-formed execution trajectories, enhancing decision-making.

Moreover, many existing datasets, although large in scale, fail to fully support complex reasoning tasks due to the lack of detailed explanations and reasoning processes in their annotations. To address this, we develop the VLA-CoT data engine, which generates the high-quality \textbf{VLA-CoT-13K dataset}, making reasoning steps explicit. The engine aligns CoT with affordance and trajectory annotations, encouraging the model to learn task-consistent reasoning and enabling it to acquire basic reasoning capabilities during the SFT phase.

Finally, we conduct comprehensive evaluations of VLA-R1 across in-domain, out-of-domain, simulation, and real-robot settings. Empirically, VLA-R1 achieves an IoU of 36.51 on the in-domain affordance benchmark, a 17.78\% improvement over the baseline; on the in-domain trajectory benchmark it attains an Average distance of 91.74 (lower is better), reducing the baseline by 17.25\%. It also delivers state-of-the-art (SOTA) performance in the out-of-domain setting. On physical hardware, VLA-R1 reaches 62.5\% success for affordance perception and 75\% for trajectory execution. These results demonstrate the method’s effectiveness under controlled conditions and its robustness and practicality across domains and real-world scenarios.

Contributions in our paper can be summarized in the following three folds:

\begin{itemize}
    \item We propose \textbf{VLA-R1}, a VLA foundation model that VLA foundation model that introduces an RLVR optimization scheme with carefully designed rewards (region alignment, trajectory consistency, and output formatting), augmented by GRPO, to systematically strengthen reasoning and execution robustness while reducing reliance on manual annotation. 
    \item We introduce the VLA-CoT data engine, which produces high-quality \textbf{VLA-CoT-13K} aligned with affordance and trajectory labels and incorporates verifiable rewards, explicitly remedying the lack of step-wise reasoning in existing VLA models.
    \item We comprehensively evaluate VLA-R1 on in-domain and out-of-domain datasets, in simulation, and on real-robot platforms, empirically verifying its effectiveness and cross-domain generalization.
\end{itemize}

\begin{table}[t]
  \centering
  \setlength{\tabcolsep}{6pt}
  \caption{\textbf{Comparison of datasets on affordance, trajectory, reasoning, scenes, and robots.} \cmark\ indicates the dataset includes that annotation; \xmark\ indicates it does not.
  “\#Scenes” counts distinct environments. “24+” for VLA-IT means at least 24 known environments
  (from BridgeData~V2) with additional RT-1 sites not consolidated. If following the official UMD
  release, set “\#Scenes” to \textbf{3}.}
  \label{tab:dataset-comparison}
  \resizebox{\columnwidth}{!}{%
    \begin{tabular}{lccccc}
      \toprule
      Dataset & \#Aff & \#Traj & \#Reasoning & \#Scenes & \#Robot \\
      \midrule
      UMD        & \cmark & \xmark & \xmark & 4    & \textemdash \\
      VAIT       & \xmark & \cmark & \xmark & \textemdash & 13 \\
      VLA-IT     & \xmark & \cmark & \cmark & 24+  & 2 \\
      ShareRobot & \cmark & \cmark & \xmark & 102  & 12 \\
      \midrule
      \textbf{VLA-CoT-13K} & \cmark & \cmark & \cmark & 102  & 12 \\
      \bottomrule
    \end{tabular}%
  }
\end{table}

\section{Related Work}

\subsection{VLA Models}

Early manipulation research often relied on state-based reinforcement learning~\cite{geng2023,andrychowicz2020}, but these methods struggled with high-dimensional visual inputs. More recently, vision-centric approaches have become dominant, harnessing the reasoning capabilities of large language models (LLMs) to improve generalization~\cite{brohan2023a,wan2023,wang2023,li2024b,liu2024e}. VoxPoser~\cite{huang2023a} uses vision-language models to generate 3D value maps, enabling zero-shot trajectory planning. RoboFlamingo~\cite{li2023b} fine-tunes on manipulation datasets to perform language-conditioned tasks, while ManipLLM~\cite{li2024b} incorporates chain-of-thought reasoning to integrate object understanding, affordance perception, and pose prediction into an interpretable framework. Building on this line, OpenVLA~\cite{kim2024} and RoboMamba~\cite{liu2024e} leverage fine-grained CoT data and supervised fine-tuning for further performance gains~\cite{song2024hazards}. Other works, such as Embodied-Reasoner~\cite{zhang2025}, Cosmos-Reason1~\cite{azzolini2025}, and RoboBrain~\cite{ji2025}, focus on long-horizon reasoning, interpretability, and logical consistency in manipulation tasks. Despite progress, most approaches still depend on large-scale annotated datasets. In contrast, ManipLVM-R1 \cite{song2025maniplvm} reduces reliance on supervision by combining small amounts of labeled data with RLVR-based self-improvement, yielding robust generalization under limited supervision.

\subsection{RLHF for VLMs}

Large vision-language models (LVLMs) have demonstrated remarkable reasoning capabilities across diverse visual tasks~\cite{liu2024c,liu2024b,li2024a,bai2023,wang2024a,huang20253d,huang2025dc,wu2025stereoadapter,liu2025nav}. Recently, reinforcement learning with verifiable rewards (RLVR) has emerged as a promising way to enhance their reasoning abilities \cite{huang20253dr1,ge2025vasevqa}. For example, Vision-R1~\cite{huang2025a} leverages a cold-start math dataset and Progressive Thinking Suppression Training to achieve results comparable to much larger models without relying on human annotations. LMM-R1~\cite{peng2025} adopts a two-stage framework, first refining reasoning on textual data and then extending to multimodal and agent-based reasoning tasks. Similarly, VLM-R1~\cite{shen2025} applies an R1-style reinforcement learning approach to visual grounding, improving open-vocabulary detection and generalization. While these works highlight RLVR’s potential, their scope remains limited to non-embodied domains. To bridge this gap, ManipLVM-R1 \cite{song2025maniplvm}, adapts RLVR to robotic manipulation, enhancing both reasoning and action execution.
LVLMs have demonstrated remarkable reasoning capabilities across diverse visual tasks~\cite{liu2024c,liu2024b,li2024a,bai2023,wang2024a}. Recently, RLVR has emerged as a promising way to enhance their reasoning abilities. For example, Vision-R1~\cite{huang2025a} leverages a cold-start math dataset and Progressive Thinking Suppression Training to achieve results comparable to much larger models without relying on human annotations. LMM-R1~\cite{peng2025} adopts a two-stage framework, first refining reasoning on textual data and then extending to multimodal and agent-based reasoning tasks. Similarly, VLM-R1~\cite{shen2025} applies an R1-style reinforcement learning approach to visual grounding, improving open-vocabulary detection and generalization. While these works highlight RLVR’s potential, their scope remains limited to non-embodied domains. To bridge this gap, ManipLVM-R1 \cite{song2025maniplvm}, adapts RLVR to robotic manipulation, enhancing both reasoning and action execution.

\begin{figure}[t]    
\centering    
\includegraphics[width=\linewidth]{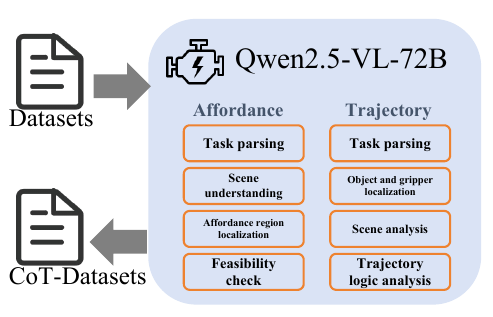}
\vspace{-2.5em}
\caption{\textbf{CoT Data Engine.} After ingesting multimodal data, the system parses tasks based on type (e.g., affordance or trajectory), performs scene understanding and localization, validates feasibility, and generates structured CoT traces for training.}
\label{fig:cot_data_engine}
\vspace{-1em}
\end{figure}

\section{Method}

\subsection{Overview}

\begin{figure*}[t]
    \centering
    \includegraphics[width=\linewidth]{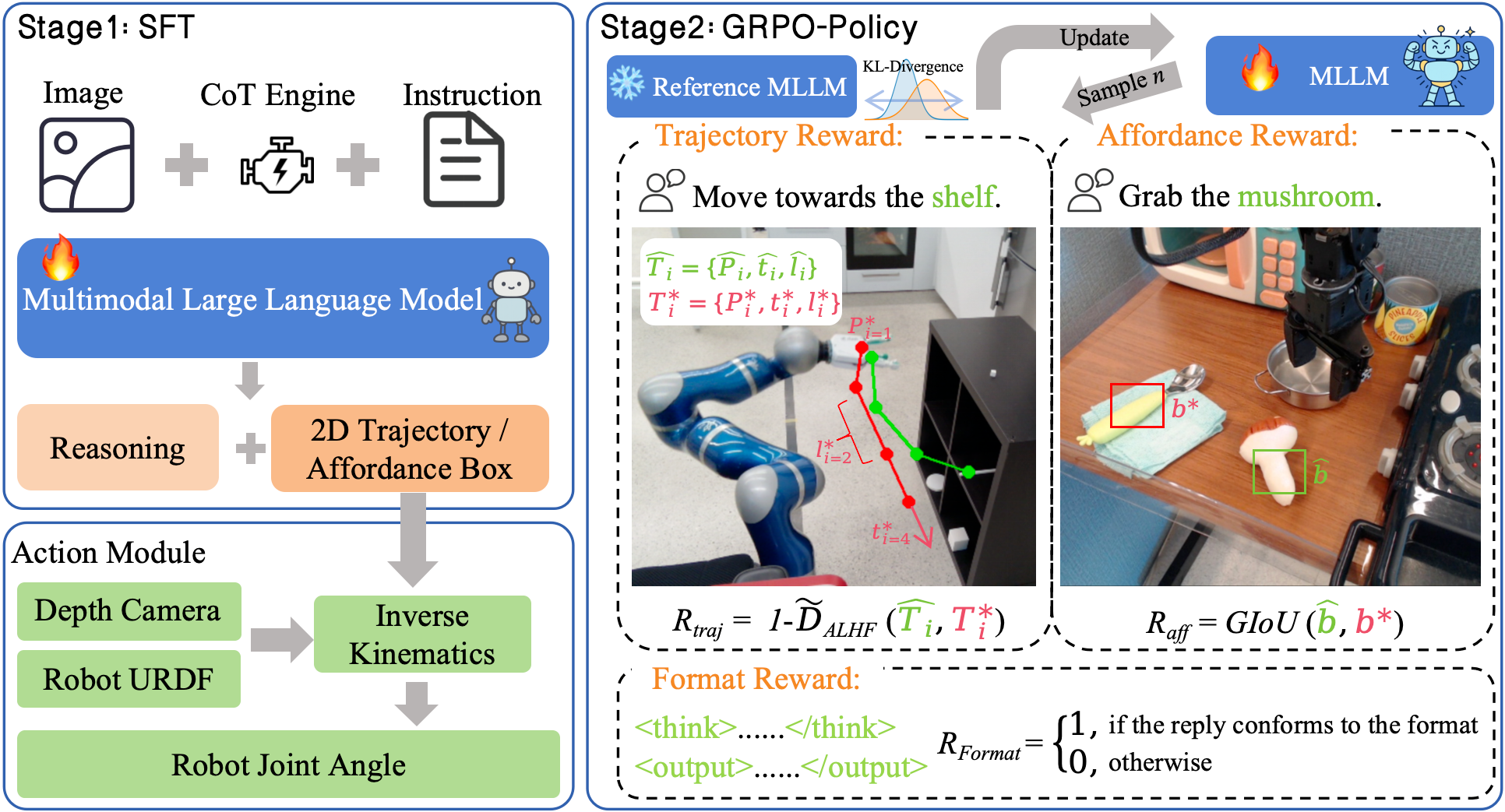}
    \vspace{-1.5em}
    \caption{\textbf{Overall architecture of VLA-R1.} 
    Training has two stages: \textbf{Stage~1} uses SFT with CoT supervision to learn reasoning over images and instructions; \textbf{Stage~2} refines reasoning and actions via RL with verifiable rewards (GRPO). 
    \textbf{During inference}, a control stack converts outputs into joint-level robot commands.}
    \label{fig:arch}
    \vspace{-1.5em}
\end{figure*}

The overall architecture of \textbf{VLA-R1} is shown in Fig.~\ref{fig:arch}. Given an input image and a natural language instruction, VLA-R1 encodes multimodal information through a vision-language backbone and then produces low-level control signals via an action decoder. Specifically, the vision branch processes raw images through a visual encoder that projects features into a shared embedding space. In parallel, the language branch tokenizes and embeds the task instruction. Both modalities are fused in the multimodal decoder, which jointly reasons over visual cues, textual context, and temporal history to generate a structured output consisting of a reasoning segment and an action prediction. The reasoning trace makes intermediate steps explicit, while the action output is represented in a discrete token space. Finally, the action de-tokenizer maps the predicted tokens into continuous 7D robot actions ($\Delta x$, $\Delta \theta$, and $\Delta$Grip), which can be directly executed on the robot arm. This design allows VLA-R1 to bridge high-level task descriptions with grounded low-level control, while maintaining interpretability through explicit reasoning traces.

\subsection{Data Synthesis}

To further strengthen the reasoning ability of our model, we construct a CoT dataset using Qwen2.5-VL-72B. As shown in Table \ref{tab:dataset-comparison} and Figure \ref{fig:cot_data_engine}, we employ the model to automatically generate intermediate reasoning steps for both affordance and trajectory tasks. In total, 13K CoT annotations are produced, which serve as high-quality supervision to bridge perception and action. These CoT data not only enrich the semantic interpretability of the training corpus but also provide explicit step-by-step guidance, enabling the model to learn task-consistent reasoning patterns.

\subsection{Supervised Fine-Tuning}

We perform supervised fine tuning on our synthetic high quality \textit{VLA-CoT-13K} dataset, which presents step by step think chains paired with grounded visual evidence and action targets. Compared with naive question and answer instruction tuning, chain of thought provides intermediate supervision signals that encourage explicit decomposition, stronger visual grounding, and stable credit assignment across time. This produces policies that reason before acting, which improves sample efficiency and prepares the model for subsequent post training under verifiable rewards. In practice we supervise both the structured \texttt{<think>} segment and the final \texttt{<output>} or action segment, which regularizes reasoning style, reduces hallucination, and yields more reliable action decoding under long horizon inputs.

We initialize our foundation model with Qwen2.5-VL-3B~\cite{bai2025qwen2}. The vision pathway is a redesigned Vision Transformer with window attention and 2D RoPE that supports native input resolution and dynamic frame rate sampling for videos. Visual tokens are softly compressed by an MLP merger before being fed into the language decoder. The text side adopts the Qwen2.5 tokenizer with a large vocabulary and the standard Qwen2.5 decoder stack. On top of the multimodal decoder we attach an action decoder that we implement to map hidden states to control outputs for downstream tasks. This initialization provides a strong balance of accuracy and efficiency for long temporal contexts.

\begin{figure*}[t]
    \centering
    \includegraphics[width=\linewidth]{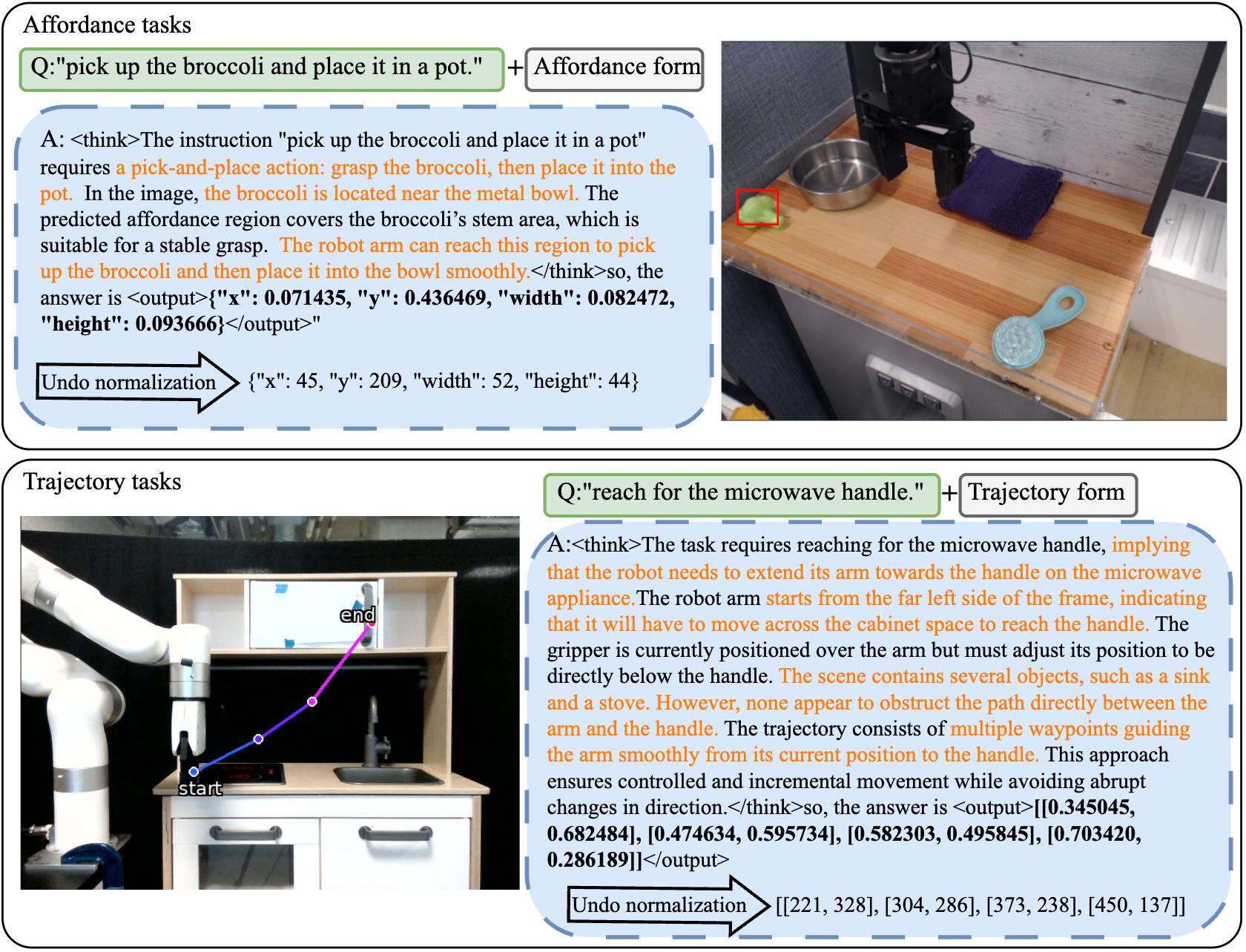}
    \caption{\textbf{Case Analysis:} The figure illustrates VLA-R1’s reasoning process and outcomes for both affordance and trajectory tasks. VLA-R1 parses the action requirements, infers relevant objects and spatial relations, and outputs the corresponding bounding boxes or waypoint sequences. The affordance form and trajectory form are fixed prompt templates that instruct the model to produce outputs in a specified format.}
    \label{fig:cot}
\end{figure*}

\subsection{Reinforcement Learning}

After SFT, we further optimize VLA-R1 through RL, as shown in Fig. \ref{fig:arch}. We adopt the GRPO algorithm, recently proposed by DeepSeek~\cite{guo2025deepseek,shao2024deepseekmath} as a scalable variant of RLHF. We extend this approach to multimodal action reasoning, allowing the model to benefit from structured verifiable rewards while maintaining training stability. For input \( q \), GRPO samples \( \{o_1, \dots, o_n\} \) from \( \pi_{\text{old}} \), scores each with a reward function to get \( r_g \). Normalize via intra-group mean \( \bar{r} \) and std \( \sigma_r \): \( \hat{A}_g = (r_g - \bar{r}) / \sigma_r \). For process supervision, step-wise rewards are normalized similarly, with token-wise advantages accumulated and shared across outputs.
For the \( k \)-th token of the \( g \)-th output, the new/old policy probability ratio is:  
\begin{equation}
r_{g,k}(\theta) = \frac{\pi_\theta(o_{g,k} \mid q, o_{g,<k})}{\pi_{\text{old}}(o_{g,k} \mid q, o_{g,<k})}.
\end{equation}

GRPO’s objective:  
\begin{equation}
\label{eq:grpo-loss}
\begin{aligned}
\mathcal{L}_{\text{GRPO}}(\theta)
={} & -\sum_{g=1}^{n} \frac{1}{|o_g|} \sum_{k=1}^{|o_g|}
\Bigl[
\min\Bigl(
r_{g,k}(\theta)\,\hat{A}_{g,k},\\[0.2em]
&\quad 
\text{clip}\bigl(r_{g,k}(\theta),\,1-\varepsilon,\,1+\varepsilon\bigr)\,\hat{A}_{g,k}
\Bigr)\;\\
&\quad
-\;\beta\,D_{\text{KL}}\bigl(\pi_\theta\parallel\pi_{\text{ref}}\bigr)
\Bigr].
\end{aligned}
\end{equation}

where \( \text{clip}(\cdot) \) bounds the ratio to \( [1 - \varepsilon, 1 + \varepsilon] \), and the last term is a KL penalty to avoid excessive policy drift.

\noindent\textbf{Fr\'echet Trajectory Reward.}
The primary reward measures alignment using  Angle-Length Augmented Fréchet distance (ALAF). Unlike pointwise Euclidean losses, ALAF respects the temporal ordering of the curves and augments it with local geometry. We represent each trajectory as a sequence of triples  $T=\{p_i, t_i, \ell_i\}$, where $p_i$ is the 2D waypoint (normalized image coordinates), $t_i$ is the \emph{unit} motion direction at $p_i$ (forward/backward difference at endpoints and a normalized blend of adjacent segment directions for interior vertices), and $\ell_i$ is the local segment length (distance to the next waypoint; for the last vertex, to the previous one). ALAF combines the positional discrete Fr\'echet term with an \emph{angle} penalty between unit tangents and a \emph{scale} penalty based on the log ratio of neighboring segment lengths, weighted by $\lambda_{\theta}$ and $\lambda_{r}$; see Eq.~\eqref{eq:osf}.

\vspace{-0.5cm}
\begin{equation}
\label{eq:osf}
\begin{aligned}
D_{\mathrm{ALHF}}(\hat T,T^*)
&= \min_{\Phi}\ \max_{(i,j)\in\Phi}\Big[
\underbrace{\|\hat p_i - p^{*}_j\|_2}_{\text{position}} \\
&\hspace{-4.1em}
+ \lambda_\theta\,\underbrace{\arccos\!\Big(\tfrac{\hat t_it^{*}_j}{\|\hat t_i\|\,\|t^{*}_j\|}\Big)}_{\text{angle}} 
+ \lambda_r\,\underbrace{\big|\log(\hat \ell_i/\ell^{*}_j)\big|}_{\text{length ratio}}
\Big],
\end{aligned}
\end{equation}
where $\Phi$ denotes all order-preserving couplings between the sequences. $\hat T=\{\hat p_i, \hat t_i, \hat \ell_i\}$ denotes the ground-truth trajectory. $T^*=\{p_i^*, t_i^*, \ell_i^*\}$denotes the predicted trajectory. The trajectory reward is defined as
\begin{equation}\label{eq:traj-reward}
R_{\text{traj}} = 1 - \tilde D_{\mathrm{ALAF}}(\hat T, T^*).
\vspace{-0.1cm}
\end{equation}
Here, $\tilde D_{\mathrm{ALAF}}$ denotes the ALAF distance normalized to $[0,1]$; larger $R_{\text{traj}}$ indicates better alignment.

\begin{figure*}[t]
    \centering
    \includegraphics[width=\linewidth]{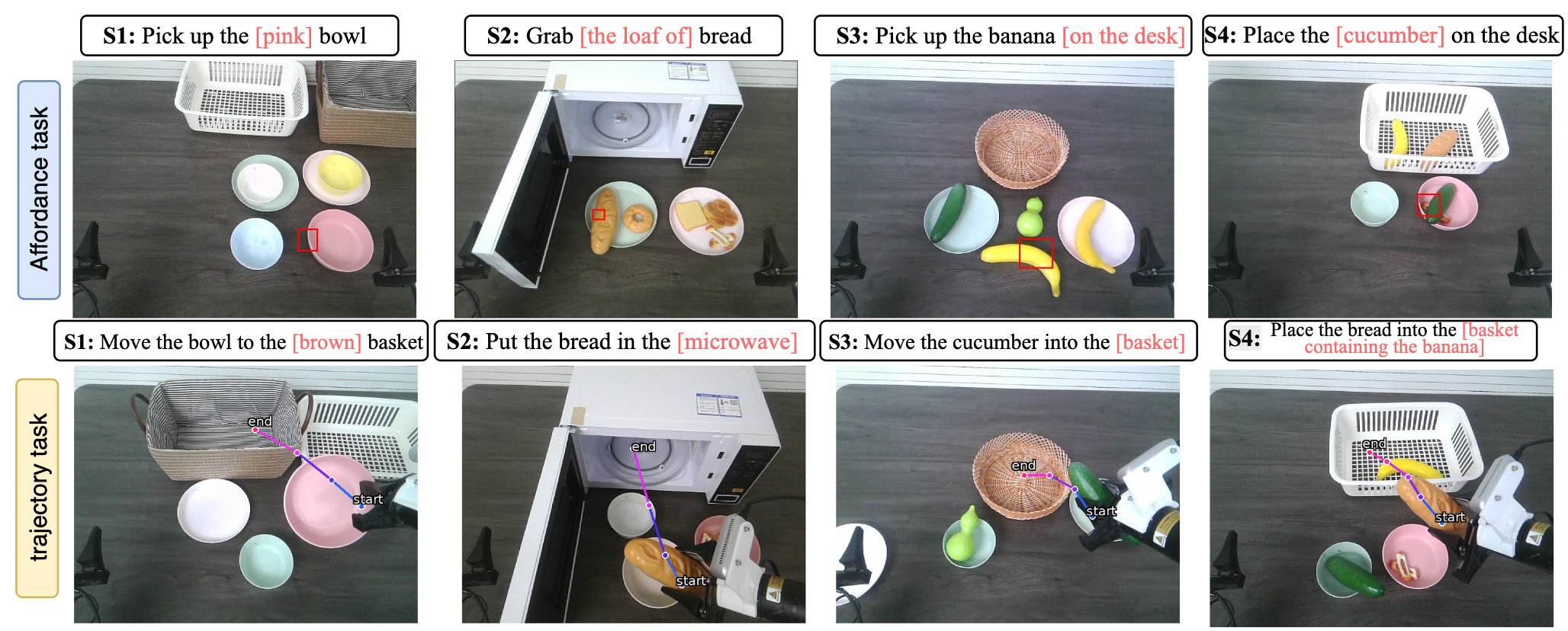}
    \caption{Visualization of evaluation in real-world scenarios.}
    \label{fig:scene}
\end{figure*}

\noindent\textbf{GIoU Affordance Reward.}
For spatial grounding, we introduce a GIoU \cite{rezatofighi2019generalized} reward between predicted and ground-truth bounding boxes. While IoU only considers the overlapping region, GIoU additionally accounts for the smallest enclosing box, penalizing misaligned predictions even when boxes do not overlap. This improves spatial robustness, especially in cluttered environments where partial overlaps are common:
\begin{equation}
R_{\text{GIoU}} = \text{GIoU}(\hat b, b^*).
\end{equation}

\noindent\textbf{Format Reward.}
Finally, we enforce structural correctness with a format reward. The model must output responses that follow the required structure (\texttt{<think>...</think>} reasoning segment followed by a \texttt{<output>...</output>} action segment). The format reward is binary:
\begin{equation}
R_{\text{format}} = 
\begin{cases}
1 & \text{if the output adheres to format}, \\
0 & \text{otherwise}.
\end{cases}
\end{equation}
This encourages interpretable reasoning traces and prevents degenerate outputs during post-training.

\section{Experiment}

To rigorously evaluate the effectiveness and generalization capacity of the proposed approach, we conduct experiments across 4 settings: in-domain datasets, out-of-domain datasets, simulation environments, and real-robot platforms. We compare with strong baselines and ablate each component to show its impact.

\subsection{Dataset and Metrics}

\subsubsection{In domain datasets} All baseline models and our proposed VLA-R1 are trained on the ShareRobot dataset\cite{ji2025}, a large-scale corpus purpose-built to advance affordance perception and trajectory prediction. ShareRobot is curated from 23 selected datasets within Open X-Embodiment\cite{o2024open}, spanning 12 robotic embodiments, 102 manipulation scenarios, and hundreds of primitive actions; it undergoes multiple rounds of human auditing to ensure high image resolution, successful task execution, accurate annotations, and complete, clean trajectory traces. The corpus comprises over one million planning question–answer pairs, 6,522 images with affordance annotations, and 6,870 images with trajectory annotations. In our experiments, we restrict training to the affordance- and trajectory-annotated image subsets and, on this basis, synthesize CoT rationales for model training.

\subsubsection{Out of domain datasets} To assess generalization, we conduct out-of-domain (OOD) evaluations. For affordance perception, we adopt a subset of the UMD Part Affordance dataset\cite{myers2015affordance} as the OOD benchmark. UMD spans 105 tools commonly encountered in gardening, kitchen, and workshop contexts. We construct our OOD test set by randomly sampling 1,200 examples from four affordance categories—grasp, cut, pound, and scoop. For trajectory prediction, we evaluate on VAIT, the validation split of LLARVA’s pretraining corpus\cite{niu2024llarva}. As VAIT originates from the highly diverse Open X-Embodiment collection, we select 500 samples and manually rectify trajectories exhibiting excessive deviation to ensure a fair and meaningful evaluation.

\subsubsection{Metric Setting}

For affordance perception, we adopt Intersection over Union (IoU) as the principal metric. IoU quantifies spatial localization fidelity by measuring the overlap between predicted and ground-truth regions; higher values indicate more accurate detection and alignment. For trajectory prediction, we evaluate the concordance between predicted and ground-truth trajectories. Following prior work\cite{ji2025,song2025maniplvm}, a trajectory is represented as an ordered set of 2D waypoints normalized to the interval [0, 1000). Similarity is assessed using three complementary metrics: Discrete Fréchet Distance (DFD), capturing global shape and temporal alignment; Hausdorff Distance (HD), measuring the maximum pointwise deviation; and Root Mean Square Error (RMSE), quantifying the average pointwise error. Together, these metrics furnish a comprehensive assessment across global shape, worst-case discrepancy, and average error, characterizing both the accuracy and consistency of trajectory prediction.

In both real-world and simulated evaluations, we report Success Rate (SR) as the task-level metric, defined as the ratio of successful executions to total trials. Success is determined as follows:  Affordance tasks: a trial is deemed successful if an object is present in the scene, the predicted bounding box correctly localizes the target object, and the system successfully grasps it; if no object is present, the model should emit no bounding box, which is likewise counted as success. Trajectory tasks: a trial is deemed successful if the executed trajectory terminates within the designated goal location (or region) and the target object is reliably transported to that endpoint.

\begin{figure}[t]
    \centering
    \includegraphics[width=\linewidth]{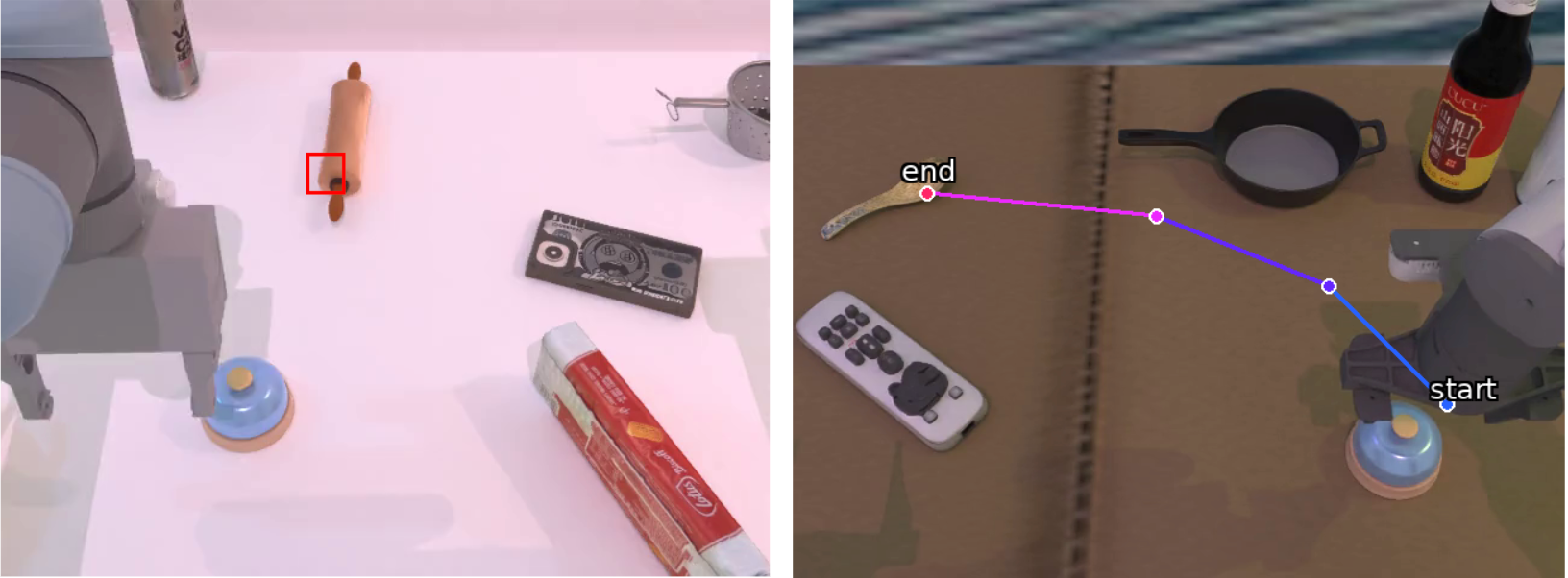}
    \caption{Visualization of simulation.}
    \label{fig:sim}
\end{figure}

\subsection{Experiment on Benchmark}

\noindent\textbf{Implementation Details.}
To ensure a fair comparison, we curate a contemporary suite of baselines. Specifically, we evaluate Gemma-3-12B-it, Gemma-3-27B-it, Phi-4-multimodal-Instruct, and the Qwen2.5-VL-3B-Instruct, Qwen2.5-VL-32B-Instruct. All open-source models are assessed under few-shot prompting to furnish a minimal perception prior. To validate the effectiveness of our training framework, we further include supervised fine-tuning baselines—InternVL2-2B , LLaVA-1.6-7B, RoboBrain-7B , and NORA-3B—as well as an RL post-trained model, ManipLVM-R1-3B.

\noindent\textbf{Experiment Results.}
As shown in the table \ref{tab:comp_benchmark}, open-source multimodal instruction-following models perform poorly on the in-domain dataset: despite large parameter counts, IoU remains below 10, and trajectory errors (DFD, HD, RMSE) are uniformly high. This indicates that generic models alone are inadequate for the precision demands of embodied tasks. Supervised fine-tuning (SFT) yields clear gains—e.g., RoboBrain-7B and NORA-3B attain higher IoU and lower trajectory errors than open-source baselines—yet their IoU typically remains in the 5–25 range. By contrast, VLA-R1-3B achieves the best results across all metrics: IoU = 36.51, DFD = 106.2, HD = 97.9, and RMSE = 71.12. Relative to the strong baseline ManipLVM-R1, IoU improves by ~17.78\%, and the overall trajectory error is reduced by ~17.25\%, attesting to the effectiveness of our training paradigm.From the OOD results, despite substantial distribution shifts, VLA-R1-3B remains superior on trajectory prediction: IoU increases to 33.96, while DFD, HD, and RMSE decrease to 114.3, 98.43, and 68.97, respectively—surpassing the strongest baseline, ManipLVM-R1-3B, and demonstrating strong cross-domain generalization and robustness. For the two types of tasks, the analysis process of VLA-R1 can be seen in Fig~\ref{fig:cot}.

\begin{table}[t]
\centering
\caption{Simulation Evaluation on Different Robot Platforms.}
\label{tab:comp_in_simulation}
\setlength{\tabcolsep}{6pt}
\begin{tabular}{llccc}
\toprule
\textbf{Model} & \textbf{Task} & \textbf{Piper} & \textbf{UR5} & \textbf{avg} \\
\midrule
\multirow{2}{*}{VLA-R1} 
 & affordance & 60\% & 50\% & 55\% \\
 & trajectory & 80\% & 60\% & 70\% \\
\midrule
\multirow{2}{*}{NORA} 
 & affordance & 50\% & 30\% & 40\% \\
 & trajectory & 10\% & 0\% & 5\% \\
\bottomrule
\end{tabular}
\end{table}

\begin{table*}[htbp]
\centering
\small
\setlength{\tabcolsep}{6.5pt}
\renewcommand{\arraystretch}{1.2}
\caption{In-domain and Out-of-domain performance comparison.}
\label{tab:comp_benchmark}
\resizebox{\textwidth}{!}{%
\begin{tabular}{l|ccccc|ccccc}
\toprule
\textbf{Method} 
& \multicolumn{5}{c|}{\cellcolor{headblue}\textbf{In-Domain}} 
& \multicolumn{5}{c}{\cellcolor{headblue}\textbf{Out of Domain}} \\
\cmidrule(lr){2-6}\cmidrule(lr){7-11}
& \cellcolor{ioupink}\textbf{IoU $\uparrow$}
& \cellcolor{metricgreen}\textbf{DFD $\downarrow$}
& \cellcolor{metricgreen}\textbf{HD $\downarrow$}
& \cellcolor{metricgreen}\textbf{RMSE $\downarrow$}
& \cellcolor{metricgreen}\textbf{Avg $\downarrow$}
& \cellcolor{ioupink}\textbf{IoU $\uparrow$}
& \cellcolor{metricgreen}\textbf{DFD $\downarrow$}
& \cellcolor{metricgreen}\textbf{HD $\downarrow$}
& \cellcolor{metricgreen}\textbf{RMSE $\downarrow$}
& \cellcolor{metricgreen}\textbf{Avg $\downarrow$} \\
\midrule

\rowcolor{groupbg}\multicolumn{11}{l}{\textbf{Open-source Models}}\\
Phi-4-multimodal-Instruct & 0.58 & 243.92 & 224.73 & 189.27 & 228.21 & 2.17 & 240.18 & 235.44 & 202.69 & 226.10 \\
Gemma-3-12b-it            & 1.18 & 206.72 & 190.64 & 154.96 & 184.10 & 4.65 & 204.94 & 209.88 & 175.42 & 193.75 \\
Gemma-3-27b-it            & 1.32 & 257.42 & 230.29 & 184.47 & 224.09 & 8.20 & 232.86 & 268.03 & 209.25 & 250.67 \\
Qwen2.5-VL-3B-Instruct    & 6.15 & 208.02 & 179.12 & 144.14 & 175.37 & 23.96 & 211.80 & 205.00 & 140.49 & 250.67 \\
Qwen2.5-VL-32B-Instruct   & 7.40 & 125.54 & 113.00 & 85.05 & 107.86 & 25.14 & 182.73 & 176.51 & 133.17 & 164.14 \\
\midrule

\rowcolor{groupbg}\multicolumn{11}{l}{\textbf{Supervised Fine-Tuning}}\\
LLaVA-1.6-7B              & 3.98 & 184.40 & 178.00 & 133.28 & 165.23 & 5.90 & 170.88 & 167.10 & 160.79 & 166.25 \\
InternVL2-2B              & 6.74 & 250.20 & 239.34 & 194.74 & 228.09 & 15.25 & 165.98 & 167.84 & 145.64 & 157.50 \\
RoboBrain-7B              & 11.79 & 156.10 & 136.52 & 106.71 & 133.11 & 22.00 & 220.94 & 214.14 & 173.02 & 202.70 \\
NORA-3B                   & 23.48 & 139.65 & 126.76 & 92.97  & 120.45 & 22.44 & 154.81 & 129.84 & 95.65  & 126.77 \\
\midrule
\rowcolor{groupbg}\multicolumn{11}{l}{\textbf{Supervised Fine-Tuning + Reinforcement learning}}\\
ManipLVM-R1-3B            & 31.00 & 134.18 & 111.14 & 87.28  & 110.87 & 28.00 & 146.82 & 140.52 & 108.64 & 131.99 \\
\textbf{VLA-R1-3B}        & \textbf{36.51} & \textbf{106.20} & \textbf{97.90} & \textbf{71.12} & \textbf{91.74} 
                          & \textbf{33.96} & \textbf{114.30} & \textbf{98.43} & \textbf{68.97} & \textbf{93.90} \\
\bottomrule
\end{tabular}
}
\end{table*}

\subsection{Experiment on Simulation}
\noindent\textbf{Implementation Details.}
To assess performance under controlled yet stochastic conditions, we conducted additional experiments in a simulated tabletop environment. Using the RoboTwin simulator, we instantiated a single randomized tabletop clutter generator that dynamically varies object categories, colors, poses/positions, and the table color throughout each trial. To evaluate the cross-robot generality of VLA-R1, we tested two robotic embodiments—Piper and UR5. Each embodiment was evaluated over ten independent trials with randomized initialization.

\noindent\textbf{Experiment Results.}
Because our training data are drawn entirely from real-world settings, the simulated environment exhibits greater variability; nevertheless, as shown in Table~\ref{tab:comp_in_simulation} anf Fig.~\ref{fig:sim}, VLA-R1 attains strong performance on both tasks. For affordance perception, VLA-R1 achieves 6/10 successes on Piper and 5/10 on UR5 (average SR = 55\%). For trajectory execution, performance improves to 8/10 on Piper and 6/10 on UR5 (average SR = 70\%), indicating that once a reliable grasp is established, the trajectory policy remains highly stable in simulation. By contrast, the NORA baseline performs notably worse under the same conditions: on the affordance task, SR drops to 50\% (Piper) and 30\% (UR5); on the trajectory task, it records 1/10 on Piper and 0/10 on UR5. Overall, these results confirm that VLA-R1 preserves robust cross-embodiment stability and superior generalization, even under heightened environmental variation.

\subsection{Experiment on the Real World}
\noindent\textbf{Implementation Details.}
To comprehensively assess real-world performance, we design four canonical scenarios on a tabletop platform. We instantiate: \textbf{S1: Bowl picking}, containing bowls of multiple colors placed in diverse locations; the model must grasp the user-specified color and, for trajectory tasks, place it precisely into a designated frame/basket of a given color. \textbf{S2: Fruit picking}, featuring repeated instances of the same fruit; the model must disambiguate and grasp the specified item and, for trajectory tasks, place it into the basket or onto the plate indicated by the instruction. \textbf{S3: Kitchen scenario}, comprising an open microwave, plates, and food props, where the model must contend with visual occlusion from the door and the spatial constraints of the cavity. \textbf{S4: Mixed scenario}, in which bowls, produce, baskets, and plates co-occur, requiring grasp-and-place under multi-category, multi-attribute distractors. Each scenario is evaluated over ten independent trials; we randomize initial object placements and poses and shuffle scenario order to mitigate potential ordering effects.

\noindent\textbf{Experiment Results.} As shown in Table \ref{tab:realworld_sr} and Figure \ref{fig:scene}, VLA-R1 achieves an average Success Rate (SR) of \~62.5\% across the four scenarios for affordance perception. By contrast, trajectory prediction attains a higher SR of 75\%. The NORA-3B baseline records \~35\% on affordance perception and 47.5\% on trajectory prediction. We observe that distractors such as color similarity and positional variation materially affect the model’s decisions, constituting the primary sources of error. Nevertheless, even under heavy clutter, VLA-R1’s predictions typically concentrate near the target rather than diverging arbitrarily, indicating a degree of tolerance and self-correction in perception and trajectory generation; when the target cannot be fully locked, the model still preserves reasonable local spatial consistency. Overall, these results validate the method’s practical viability in real settings and its ability to maintain stability under attribute similarity and visual clutter.

\begin{table}[t]
\centering
\caption{Real-World Experiments}
\label{tab:realworld_sr}
\setlength{\tabcolsep}{6pt}
\begin{tabular}{llccccc}
\toprule
\textbf{Model} & \textbf{Task} & \textbf{S1} & \textbf{S2} & \textbf{S3} & \textbf{S4} & \textbf{avg} \\
\midrule
\multirow{2}{*}{VLA-R1} 
 & affordance & 80\% & 60\% & 70\% & 60\% & 62.5\% \\
 & trajectory & 60\% & 80\% & 80\% & 80\% & 75\% \\
\midrule
\multirow{2}{*}{NORA} 
 & affordance & 40\% & 30\% & 30\% & 40\% & 35\% \\
 & trajectory & 40\% & 50\% & 30\% & 70\% & 47.5\% \\
\bottomrule
\end{tabular}

\end{table}

\subsection{Ablation Study}
To rigorously assess the impact of Chain-of-Thought (CoT) reasoning and Reinforcement Learning (RL) on performance, we conduct an ablation study with three configurations: (1) without CoT and RL (w/o CoT \& RL); (2) CoT only; and (3) CoT+RL. All models are trained under identical hyperparameters to ensure a fair comparison.

From the table \ref{tab:ablation}, using CoT alone—relative to the configuration without CoT or RL—raises IoU from 23.74 to 28.37 and reduces the average distance metric from 128.38 to 124.6. The improvement is more pronounced for IoU, indicating that CoT confers a degree of attribute disambiguation and thus benefits affordance-centric tasks. When combined with RL, the model achieves substantial gains across all metrics, underscoring the complementarity of CoT and RLVR in trajectory prediction: CoT provides structured task decomposition and reasoning, while RLVR leverages reward signals to refine execution policies, producing significant end-to-end performance improvements.

\begin{table}[t]
\centering
\scriptsize   %
\caption{Ablation study on the effect of CoT reasoning and RL. Higher IoU is better, lower DFD/HD/RMSE/Avg are better.}
\label{tab:ablation}
\setlength{\tabcolsep}{6pt} %
\begin{tabular}{lccccc}
\toprule
Method & IoU $\uparrow$ & DFD $\downarrow$ & HD $\downarrow$ & RMSE $\downarrow$ & Avg $\downarrow$ \\
\midrule
w/o CoT \& RL & 23.74 & 149.38 & 135.72 & 100.04 & 128.38 \\
CoT only      & 28.37 & 145.51 & 131.26 & 97.03 & 124.60 \\
CoT + RL      & \textbf{36.51} & \textbf{106.20} & \textbf{97.90} & \textbf{71.12} & \textbf{91.74} \\
\bottomrule
\end{tabular}
\end{table}

\section{Limitation and Future Work}
While VLA-R1 demonstrates strong performance across benchmarks, simulation, and real-robot settings, a key limitation is that it has not yet been developed or validated on other types of robotic platforms such as bi-manual robot arms and quadruped robot dogs. Extending VLA-R1 to these embodiments represents an important direction for future work, enabling broader applicability and testing its generalization in more diverse real-world scenarios.

\section{Conclusion}
In this work, we introduced \textbf{VLA-R1}, a reasoning-enhanced Vision--Language--Action model that integrates chain-of-thought supervision with reinforcement learning from verifiable rewards. By designing the \textbf{VLA-CoT-13K} dataset and incorporating an RLVR-based post-training strategy, VLA-R1 explicitly strengthens both step-by-step reasoning and execution robustness. Comprehensive experiments across in-domain, out-of-domain, simulation, and real-robot platforms demonstrate that VLA-R1 achieves state-of-the-art performance and superior generalization. We believe this work provides a promising step toward bridging the gap between reasoning quality and action execution in embodied AI.

\clearpage

\begin{figure*}[t] %
  \centering
  \includegraphics[width=\textwidth]{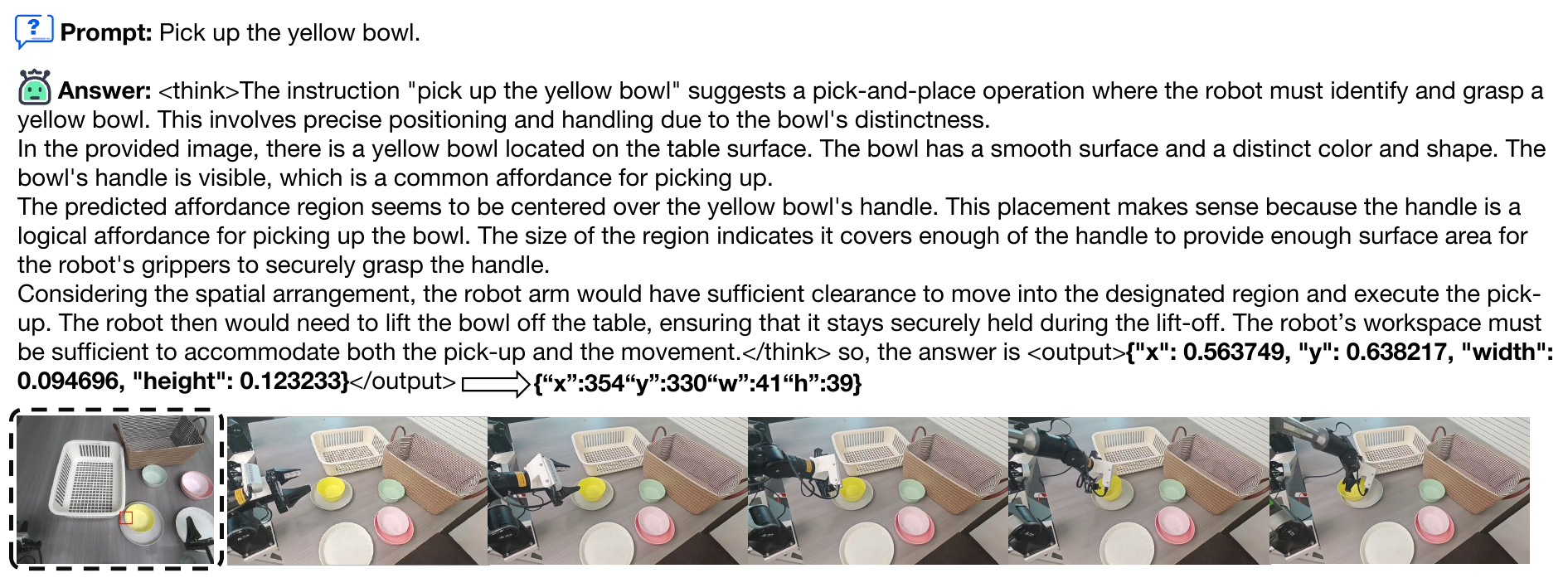}
\end{figure*}

\begin{figure*}[t] %
  \centering
  \includegraphics[width=\textwidth]{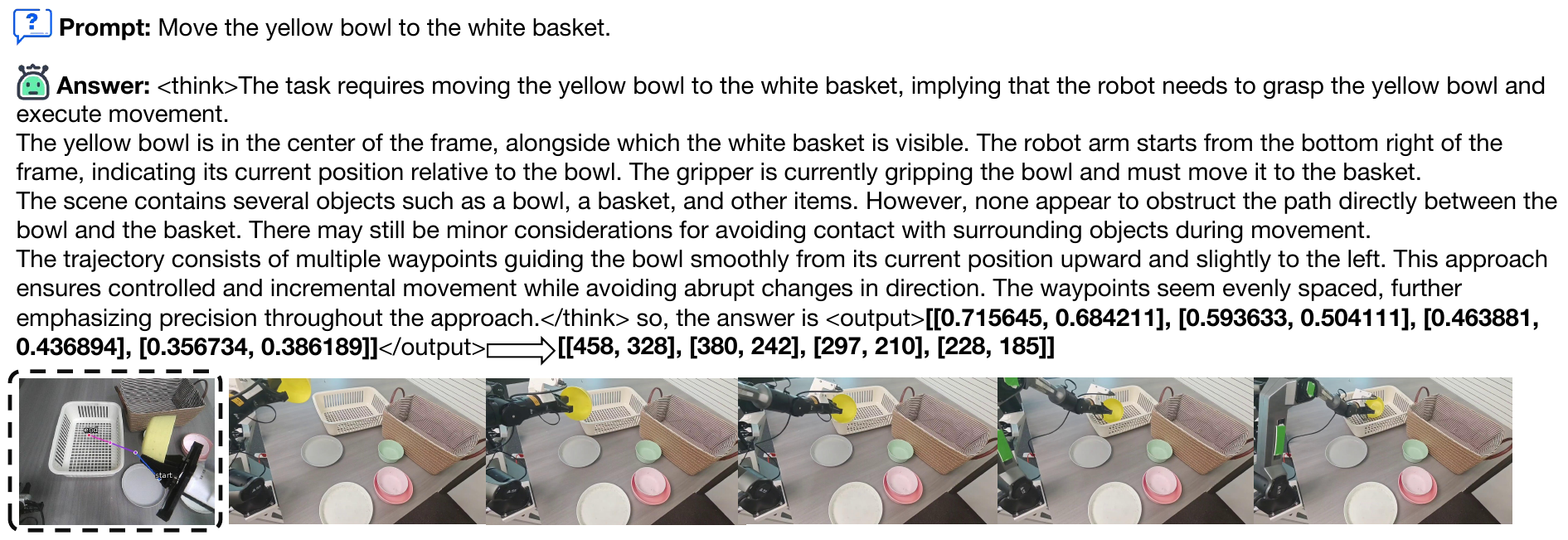}
\end{figure*}
\begin{figure*}[t] %
  \centering
  \includegraphics[width=\textwidth]{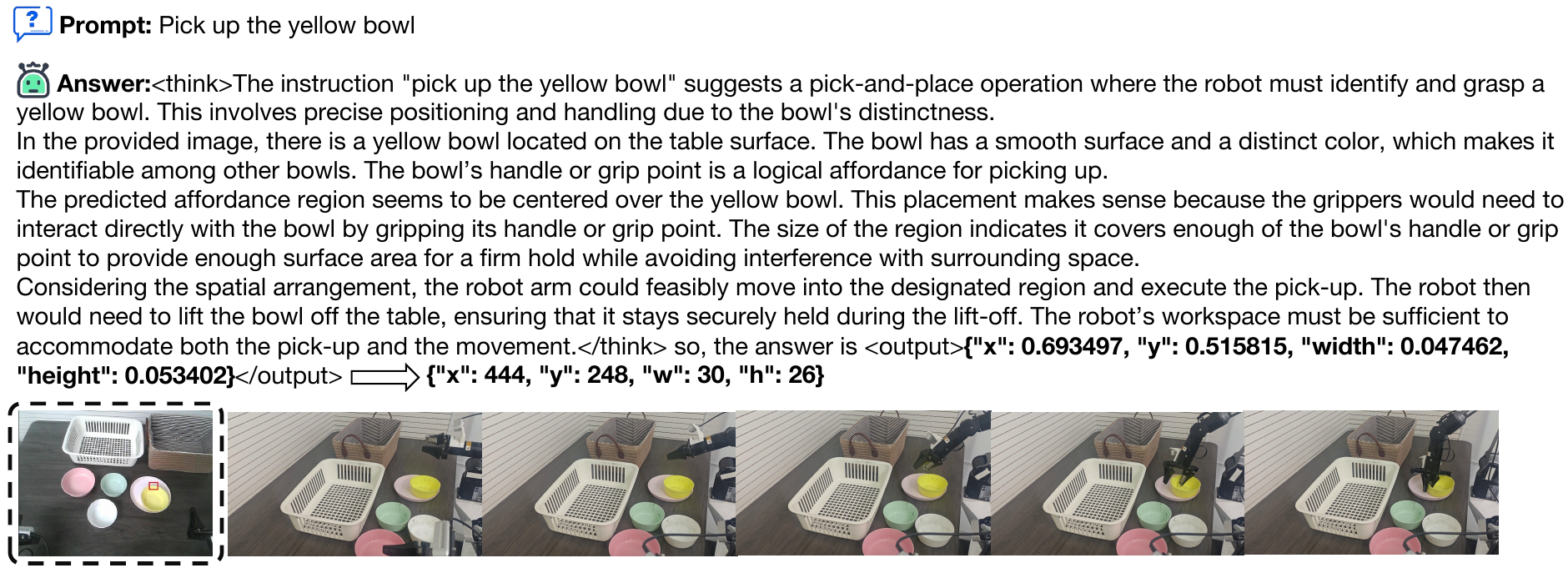}
\end{figure*}
\begin{figure*}[t] %
  \centering
  \includegraphics[width=\textwidth]{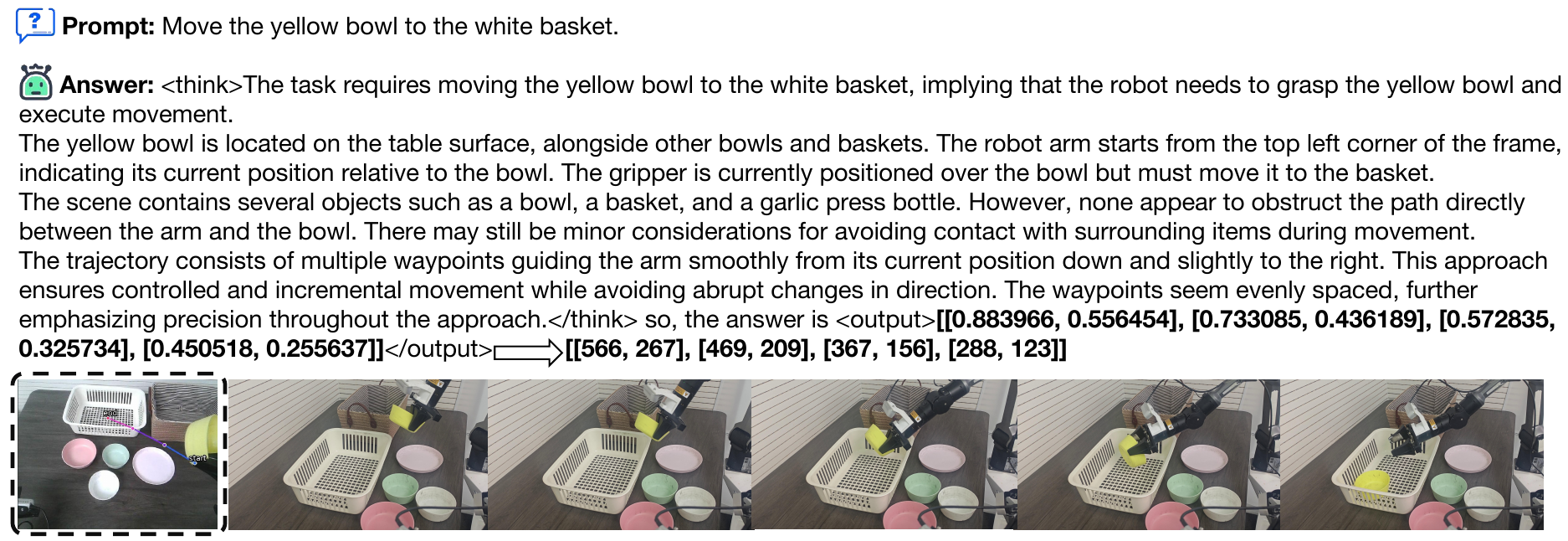}
\end{figure*}
\begin{figure*}[t] %
  \centering
  \includegraphics[width=\textwidth]{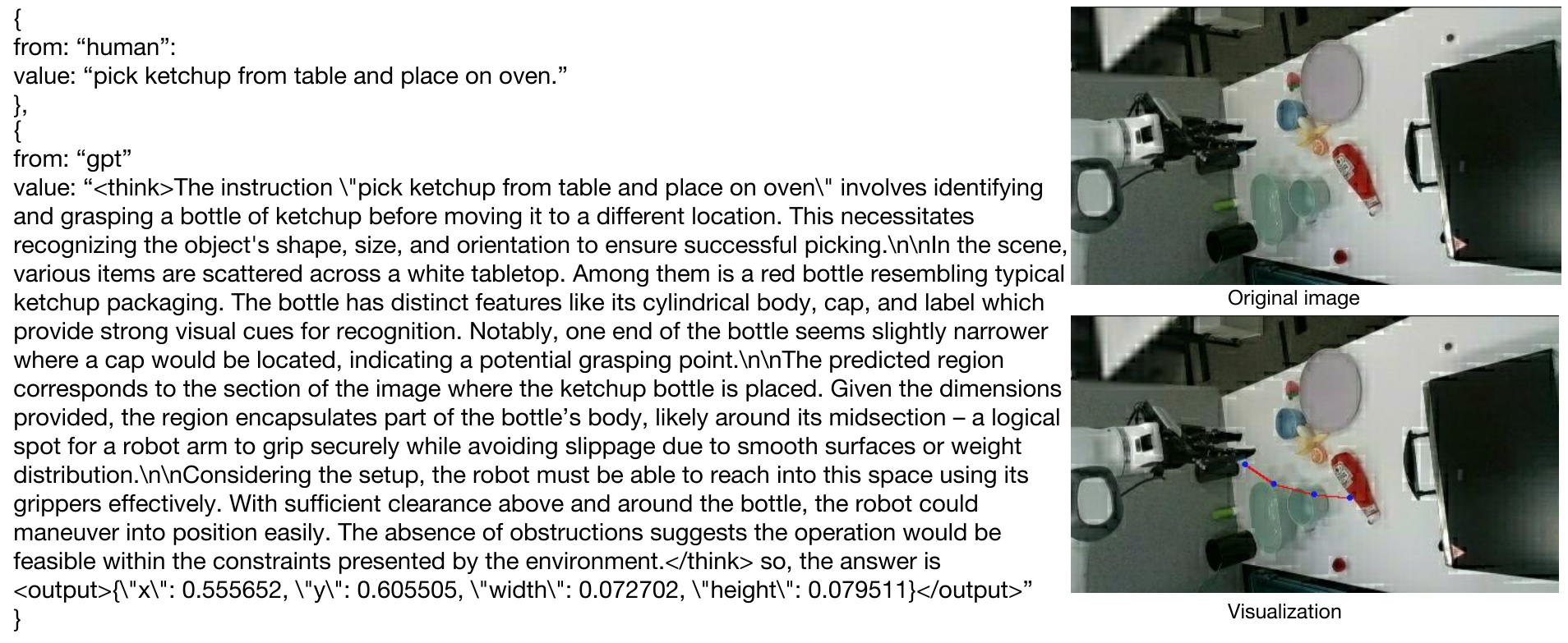}
\end{figure*}
\begin{figure*}[t] %
  \centering
  \includegraphics[width=\textwidth]{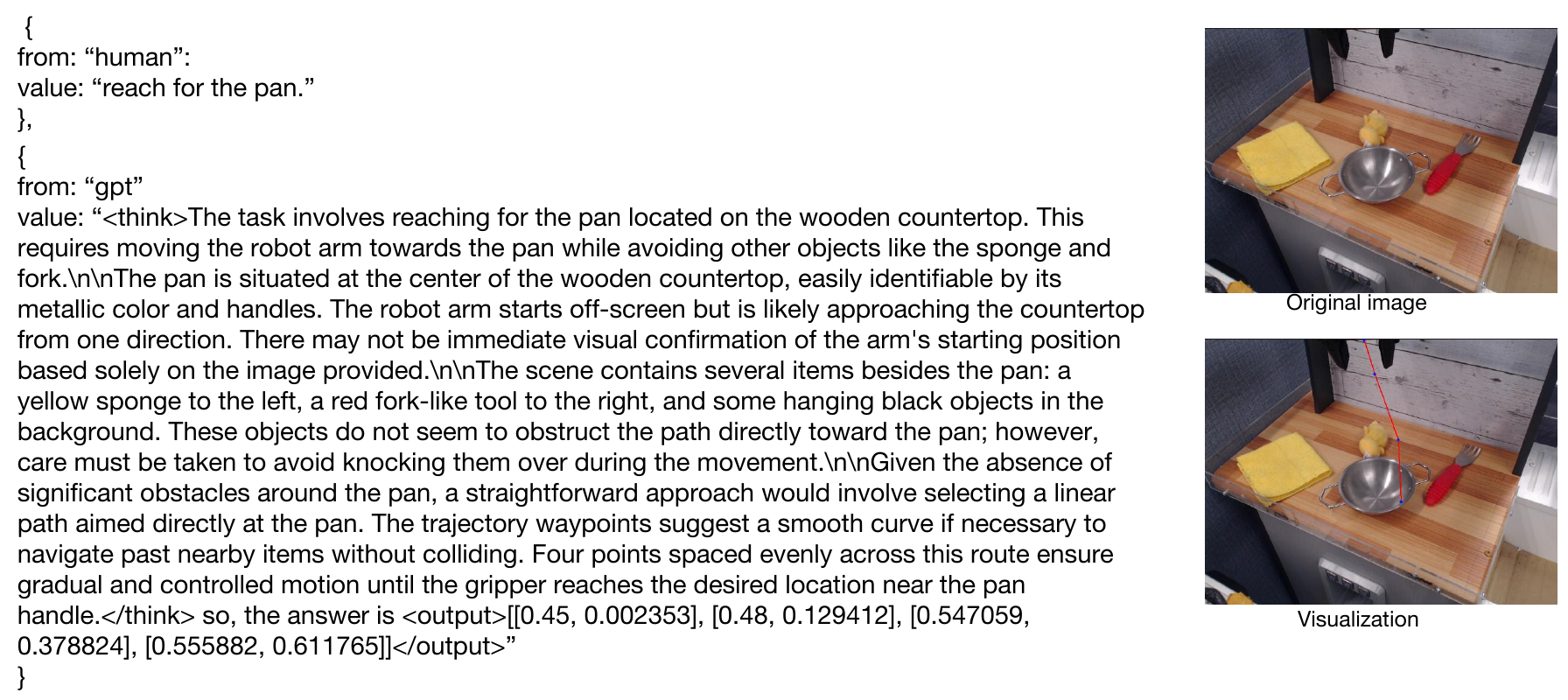}
\end{figure*}
\begin{figure*}[t] %
  \centering
  \includegraphics[width=\textwidth]{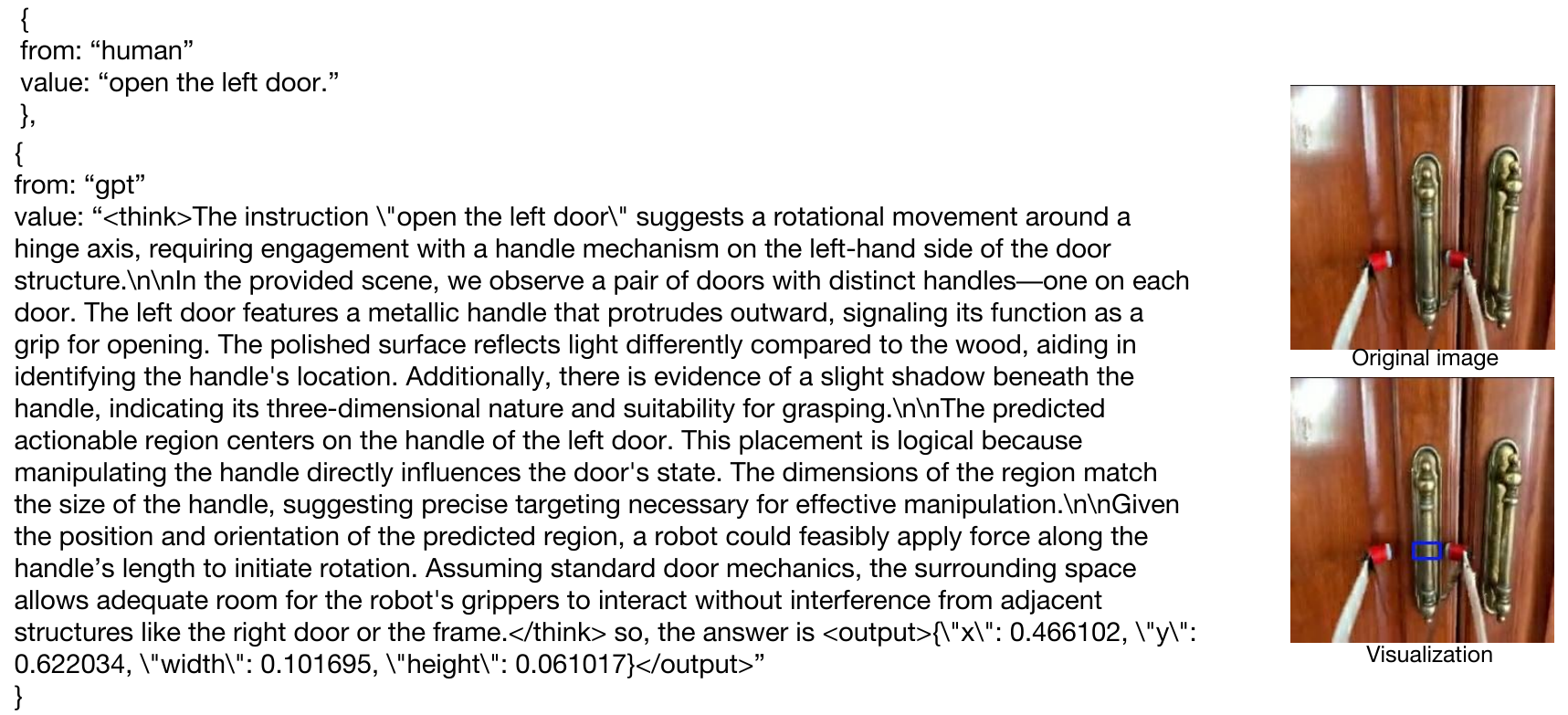}
\end{figure*}
\begin{figure*}[t] %
  \centering
  \includegraphics[width=\textwidth]{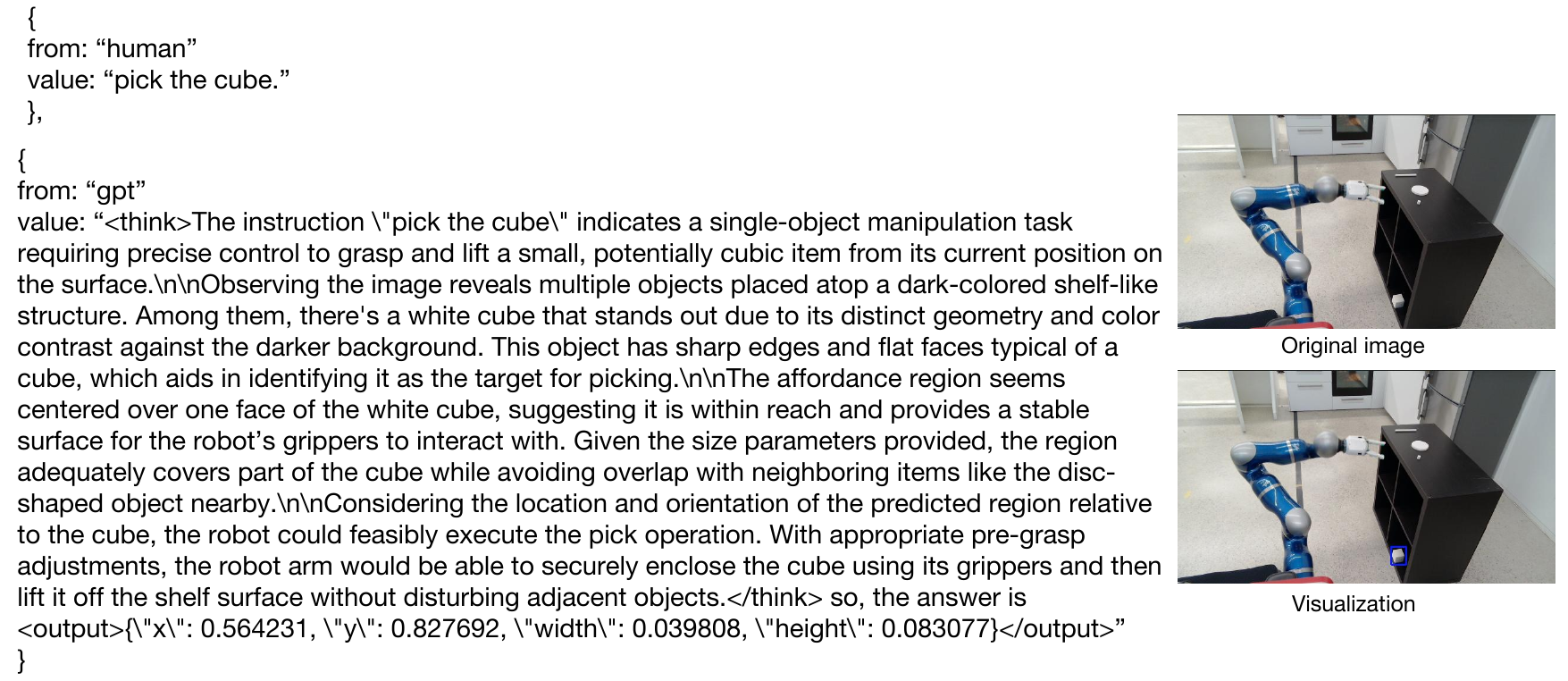}
\end{figure*}

\end{document}